\documentclass[twoside,11pt,preprint]{article} 
\usepackage{jmlr2e_preprint}

\usepackage{color}
\usepackage[utf8]{inputenc}
\usepackage[T1]{fontenc}
\usepackage{amsmath,amssymb}
\usepackage{subcaption}
\usepackage{listings}
\jmlrheading{X}{2021}{1-X}{7/21; Revised 12/21}{XX/21}{00-000}{Philipp Bach, Victor Chernozhukov, Malte S. Kurz and Martin Spindler}

\ShortHeadings{DoubleML -- Double Machine Learning in Python}{Bach, Chernozhukov, Kurz and Spindler}
\firstpageno{1}

\begin{document}

\title{DoubleML -- An Object-Oriented Implementation of Double Machine Learning in Python}

\author{\name Philipp Bach\footnotemark[2] \email philipp.bach@uni-hamburg.de \\
       \name Victor Chernozhukov\footnotemark[3] \email vchern@mit.edu \\
       \name Malte S.\ Kurz\footnotemark[2] \footnotemark[1]\phantom{\thanks{Corresponding author.}} \email malte.simon.kurz@uni-hamburg.de \\
       \name Martin Spindler\footnotemark[2] \email martin.spindler@uni-hamburg.de \\ 
       \AND
       \addr \footnotemark[2] Faculty of Business Administration,
       University of Hamburg,
       Moorweidenstraße 18, 20148 Hamburg, Germany \\
       \addr \footnotemark[3] Department of Economics and Center for Statistics and Data Science,
       Massachussets Institute of Technology,
       50 Memorial Drive, Cambridge, MA 02139, USA
       }

\editor{Joaquin Vanschoren}

\maketitle

\begin{abstract}
\texttt{DoubleML} is an open-source Python library implementing the double machine learning framework of \citet{chernozhukov2018} for a variety of causal models.
It contains functionalities for valid statistical inference on causal parameters when the estimation of nuisance parameters is based on machine learning methods.
The object-oriented implementation of \texttt{DoubleML} provides a high flexibility in terms of model specifications and makes it easily extendable.
The package is distributed under the MIT license and relies on core libraries from the scientific Python ecosystem:
\texttt{scikit-learn}, \texttt{numpy}, \texttt{pandas}, \texttt{scipy}, \texttt{statsmodels} and \texttt{joblib}.
Source code, documentation and an extensive user guide can be found at \url{https://github.com/DoubleML/doubleml-for-py} and \url{https://docs.doubleml.org}.
\end{abstract}

\begin{keywords}
Machine Learning, Causal Inference, Causal Machine Learning, Python
\end{keywords}


\section{Introduction}
The double/debiased machine learning (DML) framework \citep{chernozhukov2018} provides methods for valid statistical inference in structural equation models while exploiting the excellent prediction quality of machine learning (ML) methods for potentially high-dimensional and non-linear nuisance functions.
The Python package \texttt{DoubleML} implements DML for partially linear and interactive regression models and is primarily based on the machine learning package \texttt{scikit-learn} \citep{scikit-learn}.
A key element of the theoretical DML framework are so-called Neyman orthogonal score functions which also form the central part of the object-oriented implementation of \texttt{DoubleML}.  
The object-oriented structure also makes the package highly flexible with regards to model specifications and facilitates easy extensions.

The following sections describe the key ingredients of DML, give an introduction to the architecture of \texttt{DoubleML}, list quality standards under which the project is developed, compare the package to related software and provide some outlook to future extensions.

\section{Key Ingredients of Double Machine Learning}
Exemplarily we consider the partially linear regression (PLR) model
\begin{align}
Y = D \theta_0 + g_0(X) + \zeta, \quad &\mathbb{E}(\zeta | D,X) = 0, \label{mainreg} \\
D = m_0(X) + V, \quad &\mathbb{E}(V | X) = 0, \label{firststage}
\end{align}
where a researcher is interested in estimating the causal effect of the treatment variable, $D$, on the outcome, $Y$, provided a high-dimensional vector of confounders, $X$. 
The DML framework has three key ingredients. 

The first key ingredient is the Neyman orthogonality condition.
To estimate the causal parameter $\theta_0$, we solve the empirical analog of the moment condition
$
\mathbb{E}(\psi(W; \theta_0, \eta_0)) = 0,
$
where $W=(Y,D,X)$, $\psi(\cdot)$ is called the score function and $\eta_0=(g_0, m_0)$ are the nuisance functions.
The score function satisfies the Neyman orthogonality condition if 
$
\partial_\eta \mathbb{E}(\psi(W; \theta_0, \eta))\big|_{\eta=\eta_0} = 0,
$
where $\partial_\eta$ denotes the pathwise Gateaux derivative operator. 
Using ML estimators to approximate the nuisance functions $\eta_0$ introduces a regularization bias. 
Neyman orthogonality implies that this bias has no first-order effect on the estimation of the causal parameter $\theta_0$.

\begin{figure}[ht]
\centering
\subcaptionbox{Non-Orthogonal Score\label{fig_non_orth}}{\includegraphics[width=0.3\textwidth]{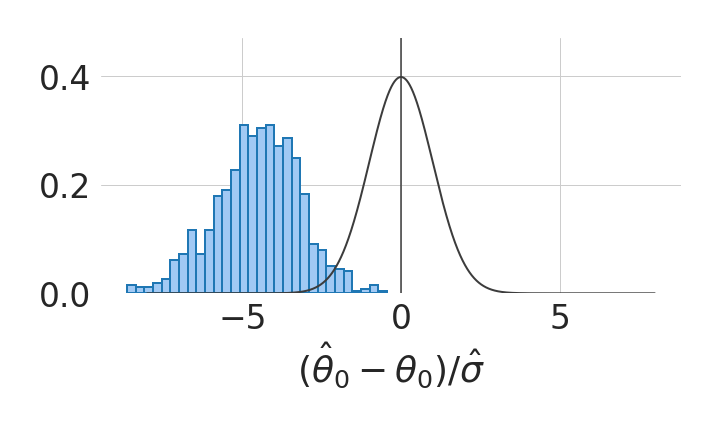}}%
\hfill
\subcaptionbox{No Sample Splitting\label{fig_no_split}}{\includegraphics[width=0.3\textwidth]{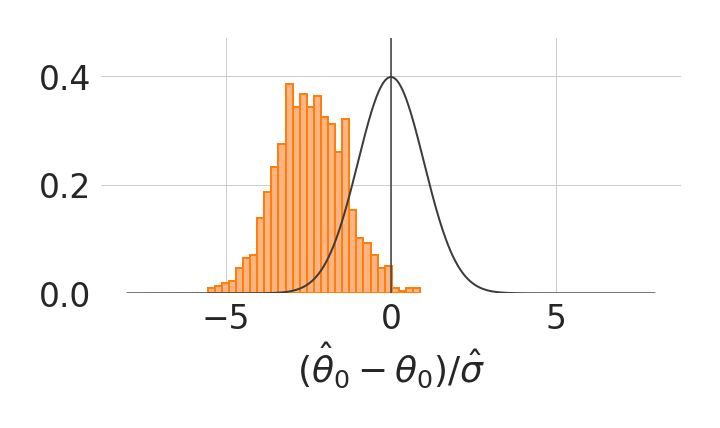}}%
\hfill
\subcaptionbox{Double Machine Learning\label{fig_dml}}{\includegraphics[width=0.3\textwidth]{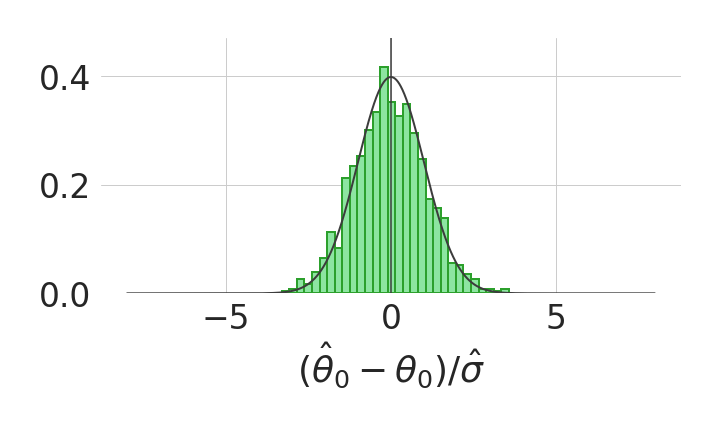}}%
\caption{The Effect of the Key Ingredients of Double Machine Learning}\label{fig_key_ingredients}
\end{figure}

To illustrate this effect, we simulate data from a PLR model. 
A naive ML approach consists of estimating $g_0$ with ML methods, for example using random forests, and then plugging-in predictions $\hat{g}_0$ to eventually obtain a naive estimate of $\theta_0$ from an OLS regression of Equation \eqref{mainreg}.  The arising bias is substantial as illustrated in Figure \ref{fig_non_orth}. 
As an alternative, we can partial out the effect of $X$ on $Y$ and $X$ on $D$ by estimating $\hat{g}_0$ and $\hat{m}_0$ with ML methods. $\theta_0$ can then be estimated from an OLS regression of $Y-\hat{g}_0(X)$ on $D - \hat{m}_0(X)$.
This approach implements a Neyman orthogonal score function that identifies $\theta_0$. As shown in Figure \ref{fig_dml}, the corresponding estimator is robust to the regularization bias.

The second key ingredient for DML are high-quality ML methods for estimating the nuisance functions $\eta_0$. It is possible to use various ML methods, including lasso, random forests, regression trees, boosting, neural nets or ensemble methods to estimate the nuisance functions.
\citet[Section 3]{chernozhukov2018} state formal conditions for the estimation quality and provide a discussion how the selection of appropriate ML methods depends on the structural assumptions for $\eta_0$, like for example sparsity.

The third key ingredient for DML is sample splitting. An overfitting bias arises if the nuisance functions are estimated on the same sample as the parameter $\theta_0$ (see Figure \ref{fig_no_split} vs.\ \ref{fig_dml}).
This bias can be overcome by sample splitting, i.e., the nuisance functions $\eta_0$ are estimated on one half of the data and the score function is solved for the target parameter $\theta_0$ on the other half of the data.
Sample splitting in $K$ folds is applicable and the usage of repeated cross-fitting is recommended to obtain more efficient estimates.

\section{Software Architecture and API}
\begin{figure}[ht]
\begin{minipage}[b]{.59\linewidth}
\includegraphics[width=0.95\textwidth]{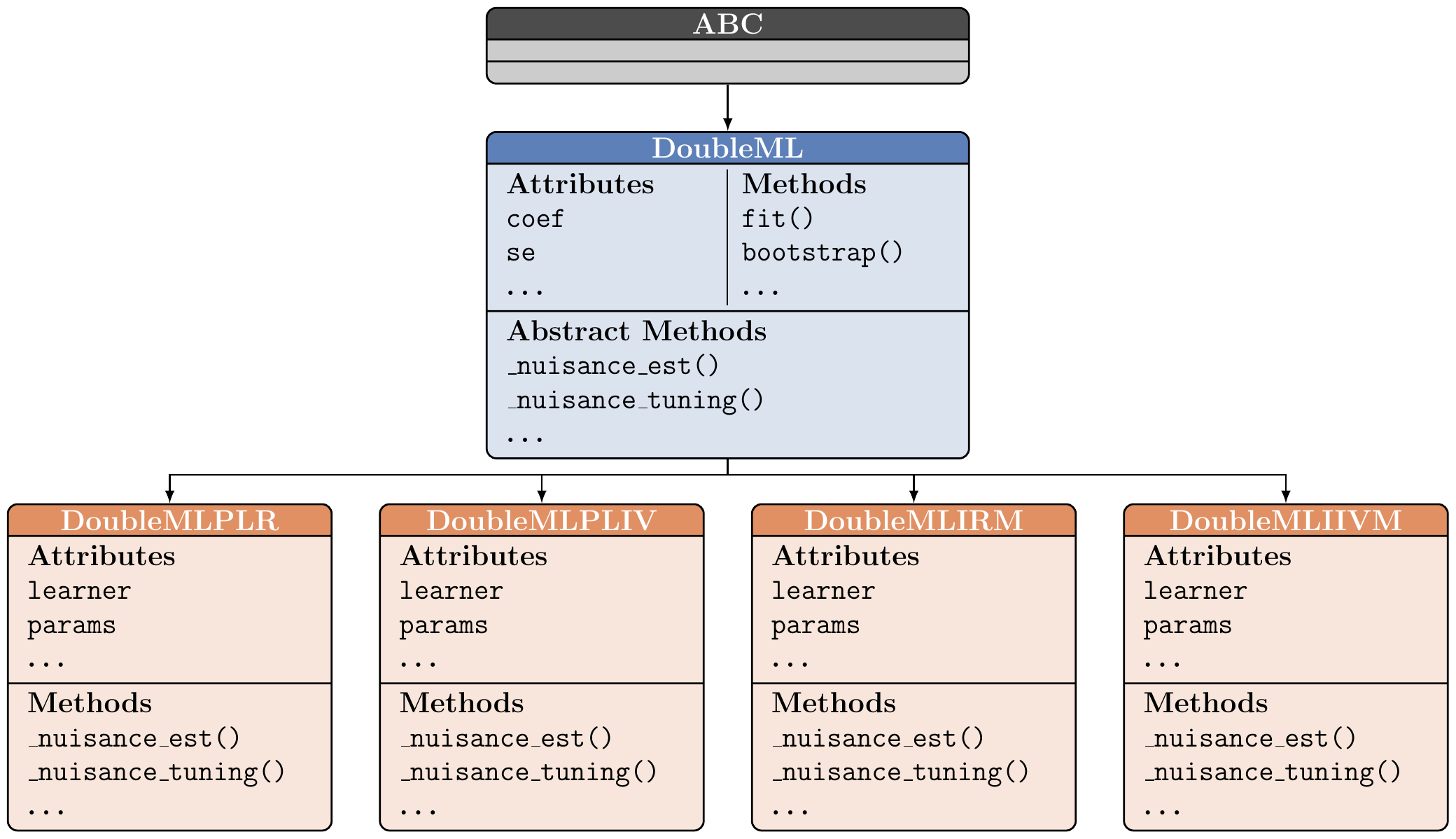}
\end{minipage}%
\begin{minipage}[b]{.415\linewidth}
\begin{lstlisting}[literate={-}{-}1,upquote=true]
import numpy as np
from doubleml import DoubleMLIRM
from doubleml import DoubleMLData
from doubleml.datasets import make_irm_data
from sklearn.ensemble import \
	RandomForestRegressor, \
	RandomForestClassifier
np.random.seed(3141)
df = make_irm_data(return_type='DataFrame',
	n_obs=1000, theta=0.5)
# DoubleMLData from a pandas.DataFrame
dml_data = DoubleMLData(df, y_col='y', d_cols='d')
dml_model = DoubleMLIRM(dml_data,
	RandomForestRegressor(max_depth=5),
	RandomForestClassifier(max_depth=5),
	score='ATE')
dml_model.fit().summary
      coef  std err        t  P>|t|  2.5 %  97.5 %
d 0.5161   0.0750 6.8805 0.0000 0.3691  0.6631
\end{lstlisting}
\end{minipage}
\caption{Class Structure and API Demonstration of the \texttt{DoubleML} Package}\label{fig_oop}
\end{figure}
Figure \ref{fig_oop} provides a summary of the object-oriented structure and a code snippet demonstrating the API of the the \texttt{DoubleML} package.
The central class of the package is the abstract base class \texttt{DoubleML} which implements all key functionalities:
The estimation of the causal parameters, standard errors, $t$-statistics and confidence intervals, but also methods for valid simultaneous inference through $p$-value adjustments and the estimation of joint confidence regions via a multiplier bootstrap approach.
The different model classes for partially linear regression, partially linear instrumental variable regression, interactive regression and interactive instrumental variable regression are inherited from \texttt{DoubleML}.
Model classes only have to add the model-specific parts which are the estimation of the nuisance models with ML methods, functionalities for the tuning of these ML methods and methods to compute the model-specific Neyman orthogonal score functions.
A key property that all mentioned model classes share is the linearity of the score functions $\psi(\cdot)$ in the parameter $\theta_0$ which is central for the object-oriented and flexible implementation of \texttt{DoubleML}.

\texttt{DoubleML} gives the user a high flexibility with regard to the specification of DML models including: The choice of ML methods for approximating the nuisance functions, different resampling schemes, selecting among the DML algorithms \texttt{dml1} and \texttt{dml2} (see \citet[Section 3]{chernozhukov2018}) and choosing from different Neyman orthogonal score functions. 
The object-oriented structure further makes it easy to extend the \texttt{DoubleML} package in different directions.
For example new model classes with appropriate Neyman orthogonal score function could be inherited from \texttt{DoubleML}, e.g., to add an implementation of DML difference-in-differences models \citep{chang2021}. 
The package features \texttt{callables} as score functions which makes it easy to extend existing model classes, e.g., the second-order orthogonal score for the PLR model by \citet{mackey2018} could be added in this way.
Additionally, the resampling schemes are customizable  in a flexible way, which for example made it possible to implement the multiway-cluster robust DML model \citep{chiang2020} with the \texttt{DoubleML} package.

\section{Development}
Releases of the \texttt{DoubleML} package are available via PyPI and conda-forge.
The source code of the package and the user guide is hosted on GitHub at \url{https://github.com/DoubleML/doubleml-for-py} and \url{https://github.com/DoubleML/doubleml-docs}. 
Collaborative work on the project is possible via the usual GitHub workflows in issues and pull requests for discussions, feature requests or bug reports.
Continuous integration with GitHub Actions ensures backward compatibility and facilitates easy integration of new code.
In the project we follow \texttt{PEP8} style standards and the package is equipped with an extensive unit test suite which leads to a 99\% test coverage.
The code quality is rated \texttt{A} by LGTM and codacy.
Documentation consists of installation instructions, a getting started manual, an extensive user guide and a detailed API reference.
It is generated with \texttt{sphinx} and hosted via GitHub Pages at \url{https://docs.doubleml.org}.
The package is distributed under the MIT license and relies on core libraries from the scientific Python ecosystem:
\texttt{scikit-learn} \citep{scikit-learn}, \texttt{numpy} \citep{numpy}, \texttt{pandas} \citep{pandas}, \texttt{scipy} \citep{scipy}, \texttt{statsmodels} \citep{statsmodels} and \texttt{joblib}.

\section{Comparison to Related Software}
The Python package \texttt{DoubleML} was developed together with an R twin \texttt{DoubleML} \citep{DoubleML2020R}.
The packages have a similar software architecture and API.
While the core functionalities are almost the same, a main difference is that the Python package is build on \texttt{scikit-learn} \citep{scikit-learn} and the R package on \texttt{mlr3} \citep{mlr3}.

Other Python packages for causal machine learning include \texttt{EconML} \citep{econml} and \texttt{CausalML} \citep{causalml}.
These packages have a focus on estimation of effect heterogeneity.
\texttt{EconML} offers a collection of different estimation methods for conditional average treatment effects, among others based on the double machine learning approach.

The object-oriented architecture of \texttt{DoubleML} is centered around orthogonal moment conditions for valid statistical inference.
The implementation closely follows the moment condition based theoretical framework in \citet{chernozhukov2018} for the estimation of causal and structural parameters like for example the average treatment effect, the average treatment effect of the treated or the local average treatment effect.
A focus of \texttt{DoubleML} is on the extensibility to new model classes which automatically inherit advanced methodology for statistical inference, like for example clustered standard errors, methods for simultaneous inference or the computationally efficient multiplier bootstrap.

\section{Future Work}
\texttt{DoubleML} is under active development and we here want to share some ideas for future extensions.
The \texttt{DoubleML} packages builds on \texttt{scikit-learn}, but we intend to design a flexible API that makes it possible to use learners from other packages like \texttt{Keras} or \texttt{XGBoost} as well.
Furthermore, there are many recently developed extensions to the DML framework \citep{chang2021, semenova2020, mackey2018, kallus2020, narita2020} that we would like to integrate into the \texttt{DoubleML} package.
Another interesting path for future research are cloud computing technologies, e.g., \cite{kurz2021} uses serverless computing for estimating DML models with the \texttt{DoubleML} package.



\acks{This work was funded by the Deutsche Forschungsgemeinschaft (DFG, German Research Foundation) – Project Number 431701914.}

\vskip 0.2in
\bibliography{doubleml_py_pkg}

\end{document}